\documentclass[letterpaper, 9 pt, conference]{ieeeconf}
\IEEEoverridecommandlockouts
\usepackage{preamble}

\title{\LARGE \bf
Reliable Robotic Task Execution in the Face of Anomalies
}

\author{Bharath Santhanam$^{\dagger}$, Alex Mitrevski$^{\ddagger}$, Santosh Thoduka$^{*}$, Sebastian Houben$^{\mathsection,*}$, and Teena Hassan$^{\mathsection}$%
\thanks{Corresponding authors: Bharath, Santhanam, Alex Mitrevski, and Santosh Thoduka}%
\thanks{$^{\dagger}$Bharath Santhanam is with NEURA Robotics, 72555 Metzingen, Germany
        {\tt\scriptsize bharath.santhanam@neura-robotics.com}} %
\thanks{$^{\ddagger}$Alex Mitrevski is with the Division for Systems and Control, Chalmers University of Technology, 41258 Gothenburg, Sweden
        {\tt\scriptsize alemitr@chalmers.se}} %
\thanks{$^{*}$Santosh Thoduka and Sebastian Houben are with the Fraunhofer Institute for Intelligent Analysis and Information Systems, 53757 Sankt Augustin, Germany
        {\tt\scriptsize <santosh.thoduka, sebastian.houben>@iais.fraunhofer.de}} %
\thanks{$^{\mathsection}$Sebastian Houben and Teena Hassan are with the Institute for Artificial Intelligence and Autonomous Systems (A$^2$S), Hochschule Bonn-Rhein-Sieg, 53757 Sankt Augustin, Germany
        {\tt\scriptsize <sebastian.houben, teena.hassan>@h-brs.de}} %
\thanks{\textcopyright 2025 IEEE.  Personal use of this material is permitted.  Permission from IEEE must be obtained for all other uses, in any current or future media, including reprinting/republishing this material for advertising or promotional purposes, creating new collective works, for resale or redistribution to servers or lists, or reuse of any copyrighted component of this work in other works.}
}

\begin{document}

\maketitle
\thispagestyle{empty}
\pagestyle{empty}


\begin{abstract}
     Learned robot policies have consistently been shown to be versatile, but they typically have no built-in mechanism for handling the complexity of open environments, making them prone to execution failures; this implies that deploying policies without the ability to recognise and react to failures may lead to unreliable and unsafe robot behaviour.
    In this paper, we present a framework that couples a learned policy with a method to \emph{detect visual anomalies} during policy deployment and to \emph{perform recovery behaviours} when necessary, thereby aiming to prevent failures.
    Specifically, we train an anomaly detection model using data collected during nominal executions of a trained policy.
    This model is then integrated into the online policy execution process, so that deviations from the nominal execution can trigger a three-level sequential recovery process that consists of
    (i) pausing the execution temporarily,
    (ii) performing a local perturbation of the robot's state, and
    (iii) resetting the robot to a safe state by sampling from a learned execution success model.
    We verify our proposed method in two different scenarios: (i) a door handle reaching task with a Kinova Gen3 arm using a policy trained in simulation and transferred to the real robot, and (ii) an object placing task with a UFactory xArm 6 using a general-purpose policy model.
    Our results show that integrating policy execution with anomaly detection and recovery increases the execution success rate in environments with various anomalies, such as trajectory deviations and adversarial human interventions.
\end{abstract}


    \section{INTRODUCTION}
    \label{sec:introduction}

    To increase the flexibility of a robot's execution and reduce the requirements on explicitly modelling the execution process, robot execution policies are often acquired using learning methods, such as imitation learning or reinforcement learning (RL).
    The usefulness and versatility of learning-based robot skills, particularly for robot manipulation, has been demonstrated in many contexts \cite{ravichandar2020,kroemer2021,tang2025}; this includes recent large robot foundation models \cite{mees2022,kim2024,openx2024,octo2024}, which are potentially able to generalise experiences between skills and even between robot modalities.

    \begin{figure}[t]
        \centering
        \includegraphics[width=\linewidth]{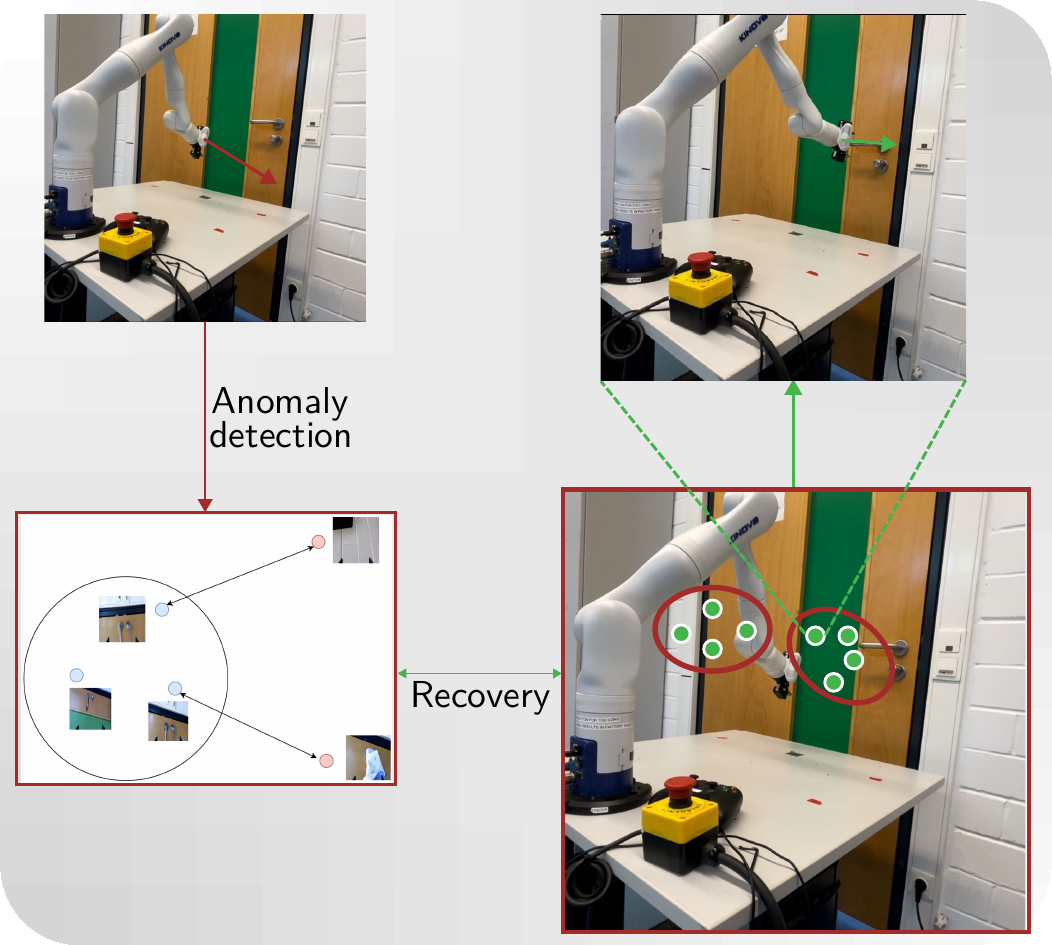}
        \caption{An overview of our proposed framework for failure-aware policy execution. During execution, visual anomalies are detected using a self-supervised feature extraction method and different recovery actions are attempted (pausing, perturbation, and finally a reset using information from a learned success model).}
        \label{fig:framework}
        \vspace{-6mm}
    \end{figure}

    Learning-based policies are, however, usually defined without considering explicit strategies for handling anomalous cases during execution; this can reduce their applicability in practical applications, as a robot is then unable to recognise and recover from situations that may lead to execution failures.
    One strategy that can be used to ensure that a policy is likely able to avoid execution failures in various scenarios (though without explicit guarantees) is domain randomisation, namely training the policy on highly diverse data; however, this poses significant requirements on the learning data quantity and variability, which is typically challenging to satisfy in practical robot applications.
    Another strategy to improve the robustness of a policy is to use model-based execution, as this would enable a robot to make predictions about the expected observations and react to discrepancies; however, versatile models of execution are generally challenging to learn, particularly over a long horizon and beyond specific modes of execution.
    This suggests that an alternative approach may be more flexible, where a robot is equipped with an ability to identify situations that may lead to execution failures and recover from them.

    In this paper, we describe such a learning-based robot system that combines learned policies with anomaly detection and execution recovery in a holistic framework.
    To detect anomalies during execution, we use self-supervised feature extraction from nominal visual data and nearest neighbour classification for identifying anomalous frames.
    In the case of anomalies, multiple recovery behaviours are attempted in succession (waiting, retrying, and then restarting the execution from a new initial state), where restarting is performed by sampling from an execution success model that is learned from the policy itself.
    The proposed framework, illustrated in Fig. \ref{fig:framework}, is agnostic of how the policy was trained.

    We experimentally verify the individual elements of the framework and the system as a whole on two tasks with two robots and policy types: (i) a door handle reaching task, where the robot follows a policy trained in simulation that is transferred to the real world, and (ii) a task of placing an object inside a bowl, for which we use a pretrained Octo model \cite{octo2024} for execution.
    The results demonstrate that, without anomaly detection and recovery, the learned policies tend to experience failures due to data distribution shifts; the additional anomaly detection and recovery increase the robustness of the policies, with the main downside that the robot may sometimes act too conservatively and initiate recovery although it could continue with its original execution.\footnote{Accompanying video: \url{https://youtu.be/oQT-xrRRvXk}}

    The paper's contributions include:
    (i) a policy-agnostic framework for monitoring execution policies and performing basic recovery,
    (ii) an evaluation of the framework's generalisation through different robots and policy types, and
    (iii) an open-source implementation of our method.\footnote{The implementation of our method can be found at \url{https://github.com/bharathsanthanam94/robust-robot-policy-execution}}

    \section{RELATED WORK}
    \label{sec:related_work}

    Robot learning is commonly used for acquiring flexible execution policies, with work often focusing on training simulated policies and transferring them to the real-world \cite{kroemer2021,tang2025,openai2021_manipulation_paper}.
    The sim-to-real transfer can introduce domain shift and lead to suboptimal policy performance, but strategies such as domain randomisation \cite{DR_main_paper} and domain adaptation \cite{RL-CycleGAN} address this challenge.
    An alternative training strategy is imitation learning on real-world data; this is followed by general-purpose robot policies \cite{kim2024,openx2024,octo2024}.
    In this work, we use both simulation-based training, combining RL with domain randomisation for sim-to-real transfer, as well as general-purpose policies trained on real-world data.

    While these methods have shown promise, the complexity of real-world environments often results in domain shifts that necessitate additional robustness measures.
    To enhance robustness, we implement visual anomaly detection, as visual data is a rich source of information about the environment and the execution itself.
    Learning approaches for anomaly detection primarily fall into two categories: reconstruction-based \cite{SSIM_AE,memorize_normality_ad} and memory bank-based methods \cite{DNN_AD,PADIM}.
    Reconstruction-based methods train models to reproduce nominal images, assuming anomalous images will be poorly reconstructed.
    In contrast, memory bank methods store features extracted from the nominal images, and classify test images as anomalous if their features are significantly different from the stored features.
    Our approach adopts the latter strategy, where we compare features of incoming images to stored image features from nominal robot task executions, due to its easier extensibility between tasks.
    In addition to pure sensor-based anomaly detection, model-based trajectory evaluation methods, such as Lagrassa et al. \cite{lagrassa2020}, can be found in the literature as well.
    We do not learn a state transition model in order to simplify the model learning requirements, particularly as we aim for a policy-agnostic model that can also be easily integrated with large-scale, model-free policies.

    Detecting anomalies during execution facilitates the implementation of recovery steps for safe execution.
    Hung et al. \cite{Uncertainty_recovery} integrate task execution with recovery using an imitation learning-based policy with concrete dropouts.
    The approach outputs a distribution of predicted actions, where the mean represents the chosen action, while high variance indicates uncertainty and triggers recovery actions.
    One limitation of \cite{Uncertainty_recovery} is that it tightly couples the task execution, monitoring, and recovery, requiring changes across all modules when modifications are made to one.
    Similarly, Lee et al. \cite{lee2019} learn a policy with uncertainty awareness so that anomalies can be detected.
    Concretely, imitation learning is used for acquiring a policy, and both epistemic and aleatoric uncertainty are estimated during execution; if the uncertainty reaches a high level, the execution is deferred to an expert policy.
    As \cite{Uncertainty_recovery}, \cite{lee2019} tightly couples the task execution and monitoring, and also uses an expert policy for recovery, but this may not be available in general.
    Thananjeyan et al. \cite{recoveryRL} propose a framework that decouples the task execution and recovery policies.
    Here, the agent first executes a primary task policy and passes its output to a safety critic, which returns the probability of the action being unsafe; if this probability is high, a recovery policy is triggered to guide the agent to a safe state.
    Similar to \cite{recoveryRL}, our approach maintains modularity by decoupling the task execution, anomaly detection, and recovery components; this property makes it applicable to various policy types (including general-purpose policies, which would otherwise have to be architecturally modified and retrained).
    Unlike \cite{recoveryRL}, we do not use a learned safety critic; training this requires data with various failure modes, which can be challenging to obtain in general.
    Reichlin et al. \cite{reichlin2022} propose a recovery method that learns a nominal state distribution density, and uses the gradient of this density to find recovery actions in out-of-distribution states.
    Similarly, Mitrano et al. \cite{mitrano2021} learn a model that evaluates whether action candidates will lead to successful recovery in a given state.
    Similar to \cite{reichlin2022}, we learn using data of successful executions, but we learn an anomaly detector rather than a recovery model.
    The recovery model in \cite{reichlin2022,mitrano2021} could be used as a substitute of the first two levels of our recovery pipeline, though at a higher learning and overall computational cost.

    Large language models (LLMs) and vision-language models (VLMs) have also been used for anomaly detection and recovery.
    Agia et al. \cite{agia2024} propose a method to monitor failures due to erratic behaviour or in task progress; for the former, the execution horizon by a diffusion policy is used for deriving an action consistency measure, while, for the latter, a VLM is queried with historical execution data and a description of the task objective.
    Liu et al. \cite{liu2023} propose a framework based on which multimodal robot data is encoded into scene graphs and used to extract relevant execution events, which are summarised in natural language and used as input to an LLM that identifies anomalies and proposes recovery plans.
    Conceptually, our anomaly detection method has similarities with the consistency estimation in \cite{agia2024}, but we estimate the consistency implicitly by comparing with successful execution examples, and we do not assume a diffusion policy, as we aim for a policy-agnostic method.
    Our detection and recovery strategy could, in principle, be integrated with a method such as \cite{liu2023}, which would be used for recovering from plan failures, while our method would be used for short-term recovery of the ongoing skill.

    \section{FAILURE-AWARE ROBOT POLICY EXECUTION}
    \label{sec:methodology}

    In this paper, we present a methodology that aims to improve the robustness of a learning-based policy through anomaly detection and subsequent recovery.
    To achieve this, we roll out an execution policy in the target environment to collect nominal execution data, which we use to train a vision-based anomaly detection model.
    During deployment, we use this model to detect anomalies at each execution step, and trigger recovery actions when necessary.
    In this section, we describe the elements of our framework in more detail.

    \subsection{Execution Policy}
    \label{sec:methodology:execution}

    As is common in the literature, we model the execution by a policy $\pi: (S, C) \rightarrow A$, which maps the robot's state $\vec{s} \in S$, and optional context $\vec{c} \in C$, to appropriate actions $\vec{a} \in A$.
    We assume a parameterised policy $\pi_\phi$, modelled by a neural network with parameters $\phi$. 
    Our monitoring and recovery strategy is independent of the nature of the policy; for concreteness, we consider two different policy types in this paper: (i) a dedicated policy learned for a concrete task, and (ii) a general-purpose, language-conditioned policy.

    \paragraph{Learning dedicated policies}
    To learn a dedicated execution policy, we use RL, so we model the execution by a Markov Decision Process (MDP); the robot thus interacts with its environment $E$ and learns to take optimal actions by maximising a cumulative reward over time.
    Concretely, at each time step $t$, the robot selects an action $\vec{a}_t = \pi_\phi(\vec{s}_t, \vec{c}_t)$, such that the environment then provides a reward $r_t = r(\vec{s}_t, \vec{a}_t, \vec{c}_t)$, which quantifies the immediate benefit of taking action $\vec{a}_t$ in state $\vec{s}_t$ under context $\vec{c}_t$.
    To ensure the applicability of the method to various robot skills and embodiments, we use a state space $S = (\mathcal{I}, Q, EE)$ that combines camera images $I \in \mathcal{I}$ as external observations, as well as proprioceptive measurements in the form of the robot's joint positions $\vec{q} \in Q$ and end effector positions $\vec{e} \in EE$.\footnote{In principle, $\vec{e}$ is not required in the input because it can be inferred from $\vec{q}$, but we use it explicitly in the state to provide richer context to the robot, thereby reducing the complexity of the learning process.}
    While the images $I$ can be from any type of camera, we use images from a wrist camera in this work. 
    We use an action space $A$ that consists of relative end effector motions $\vec{a} = (\Delta x, \Delta y, \Delta z, \Delta \alpha, \Delta \beta, \Delta \gamma)$, as task space control has been shown to speed up the learning process \cite{varin2019} and is also used in contemporary robot foundation models \cite{kim2024,openx2024,octo2024}.
    Since training an RL agent typically requires a large number of interactions with $E$, we train our policy in a simulated environment $E_{sim} \approx E$ and then transfer it to the real environment $E$.
    To ensure that the simulation-based policy is as invariant as possible to the visual domain shift that occurs due to the transfer, we use domain randomisation as a data augmentation technique.\footnote{In this work, we only use colour randomisation, as this was empirically sufficient for the task of interest.}

    \paragraph{General-purpose policies}
    Our anomaly detection and recovery method is agnostic to the nature of the policy training process, and is thus also applicable to general-purpose robot foundation models that have been trained on large imitation learning datasets, such as \cite{openx2024}.
    To demonstrate this aspect, we also consider a language-conditioned visuomotor policy $\pi_\phi^G(\vec{s}_t, \vec{c}_t)$, such that $S = \mathcal{I}$, and $C$ are language commands.
    In this case, the action space $A$ comprises relative end effector motions and a gripper command, namely $\vec{a} = (\Delta x, \Delta y, \Delta z, \Delta \alpha, \Delta \beta, \Delta \gamma, g)$, where $g \in [0, 1]$ indicates whether the gripper should be closed ($g \approx 0$) or opened ($g \approx 1$).
    Just as above, we use a wrist camera as input to $\pi_\phi^G$.
    In this work, we use a pretrained Octo \cite{octo2024} model for execution.
    As Octo uses an underlying diffusion policy \cite{chi2024}, it outputs a sequence of actions at each time step $t$; for simplicity and consistency with our dedicated policies, we only execute the first action in our evaluation.

    \subsection{Anomaly Detection}
    \label{sec:methodology:anomalies}

    \begin{figure}[t]
        \begin{subfigure}[t]{0.25\linewidth}
            \includegraphics[width=\linewidth]{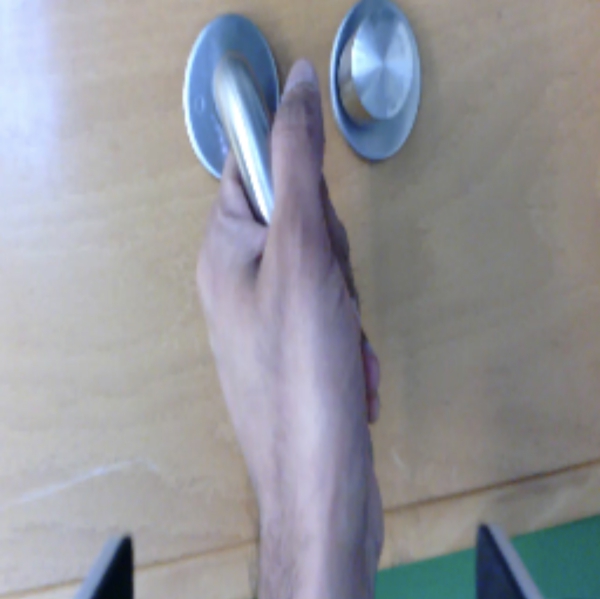}
            \caption{Handle covered by a hand}
        \end{subfigure}
        \hspace{0.09\linewidth}
        \begin{subfigure}[t]{0.25\linewidth}
            \includegraphics[width=\linewidth]{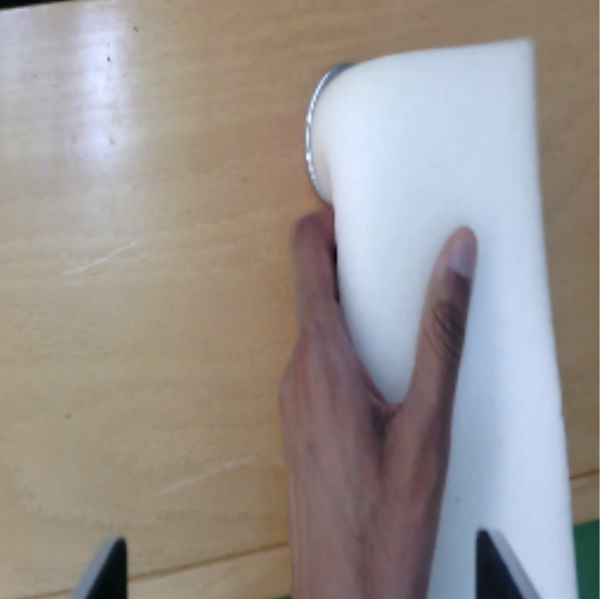}
            \caption{Handle covered by an object}
        \end{subfigure}
        \hspace{0.09\linewidth}
        \begin{subfigure}[t]{0.25\linewidth}
            \includegraphics[width=\linewidth]{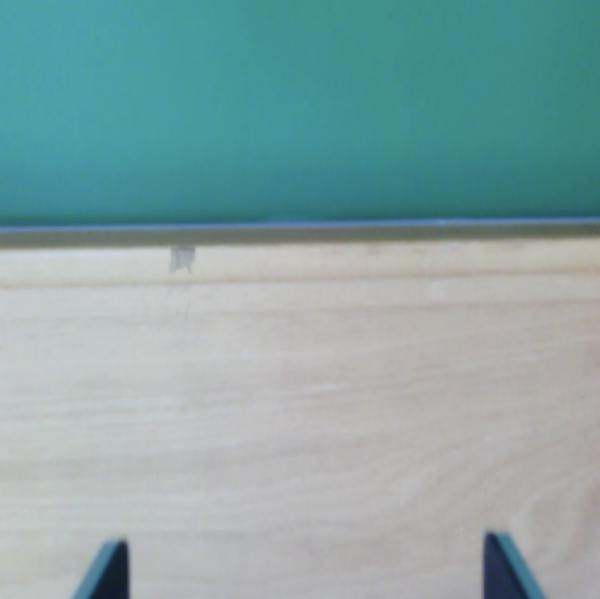}
            \caption{Deviated path}
        \end{subfigure}
        \caption{Examples of anomalies in the case of a robot attempting to reach a door handle (wrist camera view)}
        \label{fig:anomaly-examples}
        \vspace{-5mm}
    \end{figure}

    The performance of a learned policy strongly depends on the quantity and quality of the data that is used during training: a larger dataset is likely to have more diverse state and action space coverage, thereby increasing the likelihood of successful execution.
    This, however, still does not guarantee that the execution will be unaffected by anomalies, which we define as deviations from the nominal policy execution; this is because the real world is complex and can introduce many novel scenarios that the trained policy may not be equipped to handle.
    For instance, as illustrated in Fig. \ref{fig:anomaly-examples}, another agent may cover the robot's view, thereby preventing it from reaching a desired target.
    In general, it is challenging to explicitly foresee all possible anomalous scenarios during training, so a more pragmatic strategy to handle anomalies is to monitor the execution and detect \emph{any} deviations from the nominal execution as anomalies.

    Many execution anomalies can be detected using visual observations \cite{thoduka2021}; this provides rich context that can be used to detect anomalies before failures even occur.
    For instance, in Fig. \ref{fig:anomaly-examples}, if a robot detects that the target is obstructed, it can adjust its execution to avoid collisions with the external agent.
    For this reason, we propose a visual anomaly detection method that can be seamlessly integrated with the trained policy during online execution.
    The general idea is to compare online observations with observations collected during nominal policy execution, such that significant deviations are considered anomalous.
    We perform this comparison of observations at the level of image embeddings $\vec{z}$, which are extracted from a network trained in a self-supervised manner.

    The training process of the anomaly detector is summarised in Alg. \ref{alg:anomaly-detector-training}.
    Formally, we roll out the trained policy in the real world under nominal conditions of the task of interest; this results in a dataset $\mathcal{N} = \{ I_{\pi_\phi^1}, ..., I_{\pi_\phi^n} \}$ of images collected during $n$ nominal policy execution steps (line 2).
    Using these images, we fine-tune a visual feature extraction model $f_I$ trained with the DINO training scheme \cite{dino}, which uses self-distillation (line 3).\footnote{We fine-tune $f_I$ rather than use a pretrained model because the performance of anomaly detection was worse when using a pretrained model.}$^,$\footnote{Standard image augmentation, such as cropping, colour jitter, and blurring, is used for improved generalisation of the fine-tuned model.}
    Using $f_I$, we extract features from all images in $\mathcal{N}$; this results in a feature set $\mathcal{Z} = \{ \vec{z}_k = f_I(I_{\pi_\phi^k}) \; | \; 1 \leq k \leq n \}$ (lines 4--6).
    To detect anomalies during online execution, we extract features from the current image observation $I_t$, namely $\vec{z}_t = f_I(I_t)$, and estimate the distance $d$ to the nearest neighbour in $\mathcal{Z}$.
    We consider $I_t$ to be anomalous if $\underset{\vec{z}_k \in \mathcal{Z}}{\min} \; d(\vec{z}_t, \vec{z}_k) > \tau^*$, where $\tau^*$ is an anomaly threshold.
    To determine $\tau^*$, we collect an additional labelled dataset $Y$ of $m$ images with binary targets $\vec{y}$ ($0$ if fault-free, $1$ if anomalous) containing both nominal and anomalous image examples collected from $\pi_\phi$ (lines 7--8).\footnote{The threshold could, in fact, be determined based only on the statistics of the nominal data; the use of anomalous data only provides tighter constraints on the optimisation, but is in principle not required.}
    For each image in $Y$, we extract features using $f_I$ and find the distance to the nearest neighbour in $\mathcal{Z}$ (lines 9--12).
    These calculated distances $D$ are used as candidate thresholds, such that the optimal threshold $\tau^*$ is the one that leads to the highest $F$-score on $Y$ (line 13).
    \begin{algorithm}[tp]
        \begin{algorithmic}[1]
            \Function{\texttt{trainAnomalyDetector}}{$E$, $\pi_\phi$, $n$, $m$}
                \State $\mathcal{N}$ \assign \texttt{rollOutPolicySampleImgs}($\pi_\phi$, $E$, $n$)
                \State $f_I$ \assign \texttt{trainFeatureExtractor}($\mathcal{N}$)
                \State $\mathcal{Z}$ \assign $\{ \}$
                \For{$I_{\pi_\phi^i}$ \algin $\mathcal{N}$}
                    \State $\mathcal{Z}$ \assign $\mathcal{Z} \cup f_I(I_{\pi_\phi^i})$
                \EndFor
                \State $Y$ \assign \texttt{rollOutPolicySampleImgs}($\pi_\phi$, $E$, $m$)
                \State $\vec{y}$ \assign \texttt{labelNominalAnomalous}($Y$)
                \State $D$ \assign $\{ \}$
                \For{$\hat{I}_{\pi_\phi}^i$ \algin $Y$}
                    \State $\vec{z}_i$ \assign $f_I(\hat{I}_{\pi_\phi^i})$
                    \State $D$ \assign $D \cup \underset{\vec{z}_k \in \mathcal{Z}}{\min} \; d(\vec{z}_i, \vec{z}_k)$
                \EndFor
                \State $\tau^*$ \assign \texttt{findOptimalThreshold}($\vec{y}$, $D$)
                \State \Return ($\mathcal{Z}$, $\tau^*$)
            \EndFunction
        \end{algorithmic}
        \caption{Visual anomaly detection training using a self-supervised feature extractor}
        \label{alg:anomaly-detector-training}
    \end{algorithm}

    \subsection{Execution Recovery}
    \label{sec:methodology:recovery}

    The detection of anomalies is only useful if a robot can use it to trigger a recovery process that would enable it to prevent or remedy failures.
    In principle, such recovery can involve modifying the manner in which the current skill is executed or even executing a different skill, especially for long-horizon tasks \cite{mukherjee2021}.
    In this work, we only focus on recovering the currently executed skill (assuming an ergodic process \cite{moldovan2012}), with the objective of increasing the success rate of the policy executed under various anomalous situations.

    Our proposed recovery strategy consists of executing three recovery stages in succession: (i) $ER_1$: pausing the execution for a predefined number of seconds, (ii) $ER_2$: locally perturbing the robot's state, where the locality is determined by a parameter $\sigma_d$, (iii) $ER_3$: resetting the robot's initial state and retrying the policy, where the initial state is sampled from a learned success distribution $\mathcal{F}$ that is represented as a Gaussian mixture model (GMM).
    \paragraph{Pausing the execution} $ER_1$ is a useful recovery strategy when considering temporary anomalies; for example, in Fig. \ref{fig:anomaly-examples}, pausing would enable continuing the execution if the person obstructing the target leaves the robot's view.
    \paragraph{Perturbation} The assumption is that, from a state which might lead to a failure, $ER_2$ would move the robot locally to a randomised state from which the task execution can continue successfully.
    This may also overcome false positive anomaly detections: if the model outputs a false positive, pausing and retrying might give the same result, but perturbing results in a slightly different camera view, for which the anomaly detection model might give a different result.
    Providing a modified input to the model can be considered to be a variant of test-time augmentation \cite{kimura2021}.
    \paragraph{Restarting the execution} $ER_3$ abandons the current trajectory and enables the robot to retry the execution, such that, by sampling from a distribution $\mathcal{F}$ of successful starting states, the robot can be reinitialised to another state that is likely to lead to a successful execution (as illustrated in Fig. \ref{fig:framework}).
    $\mathcal{F}$ is learned in an offline learning stage from a dataset $P$ of initial states that have resulted in successful policy execution.
    This sampling strategy is inspired by the execution model proposed in \cite{mitrevski2023_ras}, but we use it without a relational model and replace the Gaussian process that is used there by a mixture model, similar to \cite{bozcuoglu2019}, for simplicity.

    The recovery stages are attempted in the above order, and the robot only proceeds to the next stage if the anomaly persists; this imposes an order based on their execution costs.
    The recovery is executed until one of the behaviours leads to success or a maximum episode length $h_{\max}$ (from the start of the execution) is reached; this means that the recovery actions are counted as execution steps, which prevents the robot from getting stuck in an infinite recovery loop.
    A summary of the policy monitoring and recovery method is given in Alg. \ref{alg:failure-aware-policy-execution}.
    \begin{algorithm}[tp]
        \begin{algorithmic}[1]
            \Function{\texttt{executePolicy}}{$E$, $\pi_\phi$, $n$, $\mathcal{Z}$, $\tau^*$, $h_{\max}$}
                \State $done$ \assign $\algfalse$, $h$ \assign $0$
                \While{\algnot $done$ \algand $h < h_{\max}$}
                    \State $h$ \assign $h$ + 1
                    \State $\vec{a}_t$ \assign $\pi_\phi(\vec{s}_t, \vec{c}_t)$
                    \State $I_{t+1}$ \assign \texttt{applyAction}($\vec{a}_t$)
                    \State $\vec{z}_{t+1}$ \assign $f_I(I_{t+1})$
                    \State $anomaly$ \assign $\underset{\vec{z}_k \in \mathcal{Z}}{\min} \; d(\vec{z}_{t+1}, \vec{z}_k) > \tau^*$
                    \If{$anomaly$}
                        \State \texttt{recover}($(ER_1, ER_2, ER_3), h$)
                    \EndIf
                    \State $done$ \assign \texttt{checkTermination}($\pi_\phi$)
                \EndWhile
            \EndFunction
        \end{algorithmic}
        \caption{Policy execution with anomaly detection and recovery. \texttt{recover} attempts the recovery behaviours described in the text in succession.}
        \label{alg:failure-aware-policy-execution}
    \end{algorithm}

    \section{EVALUATION}
    \label{sec:evaluation}

    We evaluate our integrated framework for policy execution, anomaly detection, and recovery on two different tasks and robots:
    (i) a task of reaching a door handle using a Kinova Gen3\footnote{\url{https://www.kinovarobotics.com/product/gen3-robots}} manipulator with a Robotiq 2F-85 gripper, and
    (ii) a task of putting an object in a bowl with a robot equipped with a UFactory xArm 6\footnote{\url{https://www.ufactory.cc/xarm-collaborative-robot/}} manipulator that has a parallel jaw gripper.
    In this section, we first analyse the individual elements of our framework on the door handle reaching task, and then demonstrate the generalisability of our method on the object placing task.

    \subsection{Execution Policy Learning}
    \label{sec:evaluation:execution}

\begin{figure}
        \centering
        \includegraphics[width=0.9\linewidth]{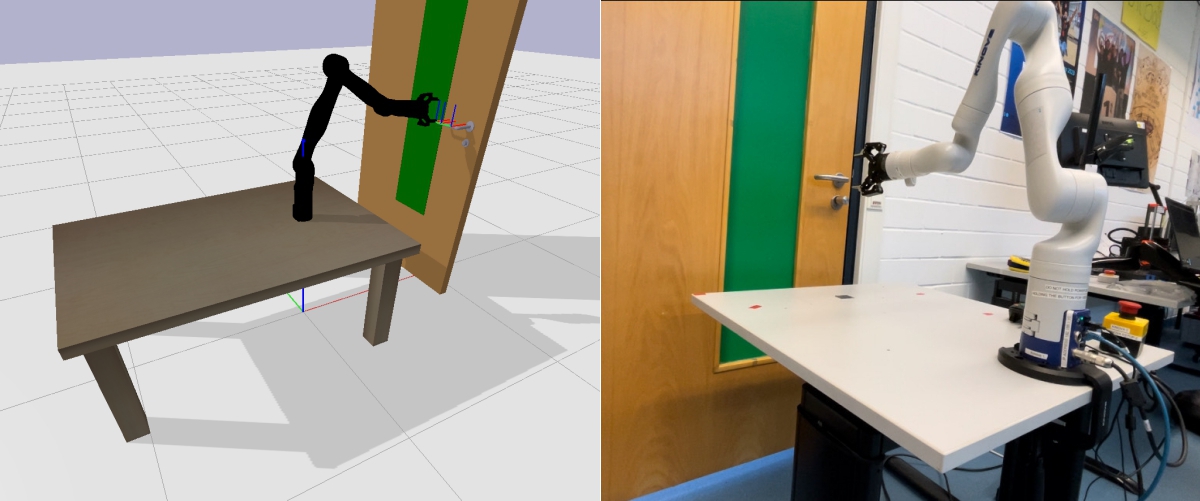}
        \caption{Experiment setup of the door handle reaching task in a simulated environment (left) and a real environment (right)}
        \label{fig:experiment-setup-kinova}
        \vspace{-0.35cm}
    \end{figure}

    For reaching a door handle, we learn a policy in a PyBullet-based\footnote{\url{https://pybullet.org/}} simulated environment that is illustrated in Fig. \ref{fig:experiment-setup-kinova}, and perform sim-to-real transfer on a door in our lab.
    During the experiment, the manipulator is fixed on a table, as illustrated in Fig. \ref{fig:experiment-setup-kinova}.
    We use the built-in RealSense D410 camera on the robot's wrist to observe the environment; examples of views from this camera are shown in Fig. \ref{fig:anomaly-examples}.
    For simplicity, the termination of the policy during the evaluation is performed manually by the experimenter.

    The policy is a multilayer perceptron that takes a 266-dimensional input that is a concatenation of (i) a 256-dimensional image feature vector extracted from a NatureCNN \cite{mnih2015} model, (ii) a 7-dimensional joint position vector, and (iii) a 3-dimensional end effector position.
    In this work, we use Proximal Policy Optimization (PPO) \cite{ppo} for learning the policy, which is an on-policy algorithm that is commonly used in the literature.
    Training is performed in an end-to-end fashion, namely the feature extractor is trained along with the policy.
    PPO trains both a policy and a value network, such that both networks have the same design, and the feature extractor is shared between both networks.

    During learning, the initial state of the robot is randomly selected within $4cm$ of a predefined point from which we could ensure that the robot has at least a partial view of the door handle.
    We use the PPO implementation in Stable Baselines3 \cite{stablebaselines3}, where we interface our simulation through a Gymnasium \cite{gymnasium} environment.
    For training, we use a shaped reward function that we define as
    \begin{equation*}
        R = w_s\mathds{1}_{\text{success}} + w_c\mathds{I}_{\text{collision}} + R_{\text{step}} + R_{\text{distance}}
    \end{equation*}
    where $w_s = 15$ (rewards the agent if it reaches the door handle), $w_c = -5$ (punishes the agent if it collides with an obstacle), $R_{\text{step}} = -0.5$ for every step (to encourage the robot to move towards the handle quickly), and $R_{\text{distance}} = \alpha\frac{d_{t-1} - d_t}{d_0}$ rewards the robot for making larger steps towards the goal; here, $d_0$ is the initial distance to the goal, $d_{t-1}$ and $d_t$ are the distances at the previous and current steps, respectively, and we use $\alpha = 10$ as a weight.
    To simplify the training process, we reduce the action space and only work with the end effector positions, but no orientations.
    Under this setup, we train the policy with a maximum allowed episode length of $100$ steps, for a total of $80{,}000$ steps, at which point the return has converged.

    The results of the evaluation of the transferred policy over $25$ trials are shown in Tab. \ref{tab:transferred-policy-success}.
    \begin{table}[t]
        \caption{Success rate of the transferred policy (out of $25$ executions) without introducing anomalies during the process}
        \label{tab:transferred-policy-success}
        \begin{tabular}{L{0.8\linewidth} | R{0.1\linewidth}}
            \cellcolor{gray!10}\textbf{No domain randomisation} & $4\%$ \\\hline
            \cellcolor{gray!10}\textbf{With domain randomisation} & $64\%$
        \end{tabular}
        \vspace{-0.5cm}
    \end{table}
    These results demonstrate that the policy can be transferred to the real environment with reasonable success (provided that domain randomisation is used during training); however, the initial position has a significant effect on the success rate and sometimes leads to unwanted deviations of the trajectory, which indicates a need for a more robust execution process.

    \subsection{Anomaly Detection}
    \label{sec:evaluation:anomalies}

    To train our anomaly detector, we collected a set of $3{,}700$ real-world image frames, which combine nominal policy executions and additional manual teleoperation examples of the manipulator for a larger variety.
    During the policy execution, the initial position of the robot was varied manually.
    For feature extraction, we fine-tune a ResNet-50 backbone \cite{resnet} (pretrained on ImageNet), using the DINO training scheme\footnote{We use the Lightly library: \url{https://github.com/lightly-ai/lightly}}, such that we have a $2048$-dimensional feature vector.\footnote{We use the trained student from DINO for feature extraction.}
    We use the Euclidean distance for detecting anomalies.

    For selecting the anomaly detection threshold $\tau^*$, we collected and labelled a dataset of $692$ frames, out of which $20\%$ were anomalous.
    The introduced anomalies were similar to those in Fig. \ref{fig:anomaly-examples}, namely they included human hand obstructions, obstructions by another object, as well as trajectory deviations.
    As alluded to by the results in Tab. \ref{tab:transferred-policy-success}, deviations sometimes occur during nominal deployment (due to observational noise), but they were also manually induced by starting the policy execution in a position where the handle is not visible.

\begin{figure}[t]
        \includegraphics[width=\linewidth]{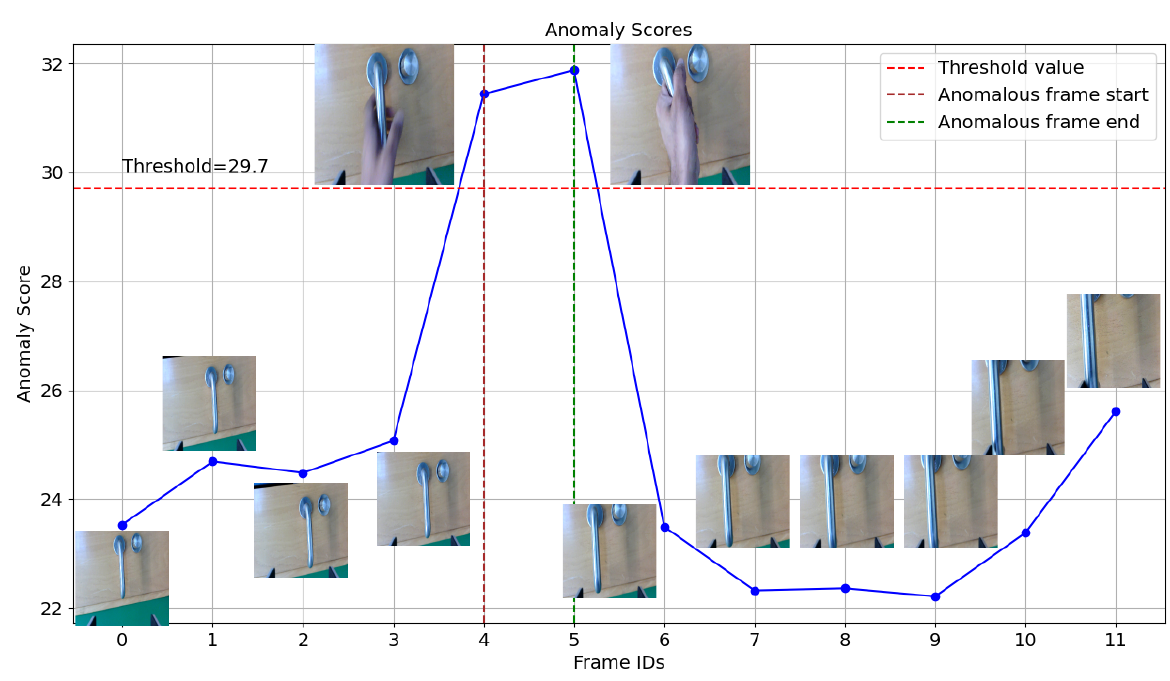}
        \caption{Illustration of the results of the anomaly detection process during policy execution in the door handle reaching task. The anomaly here is a temporary hand obstruction as illustrated in Fig. \ref{fig:anomaly-examples} (successfully detected in this case).}
        \label{fig:anomaly-detection-scores-example}
        \vspace{-4mm}
    \end{figure}

\begin{figure}[t]
        \includegraphics[width=\linewidth]{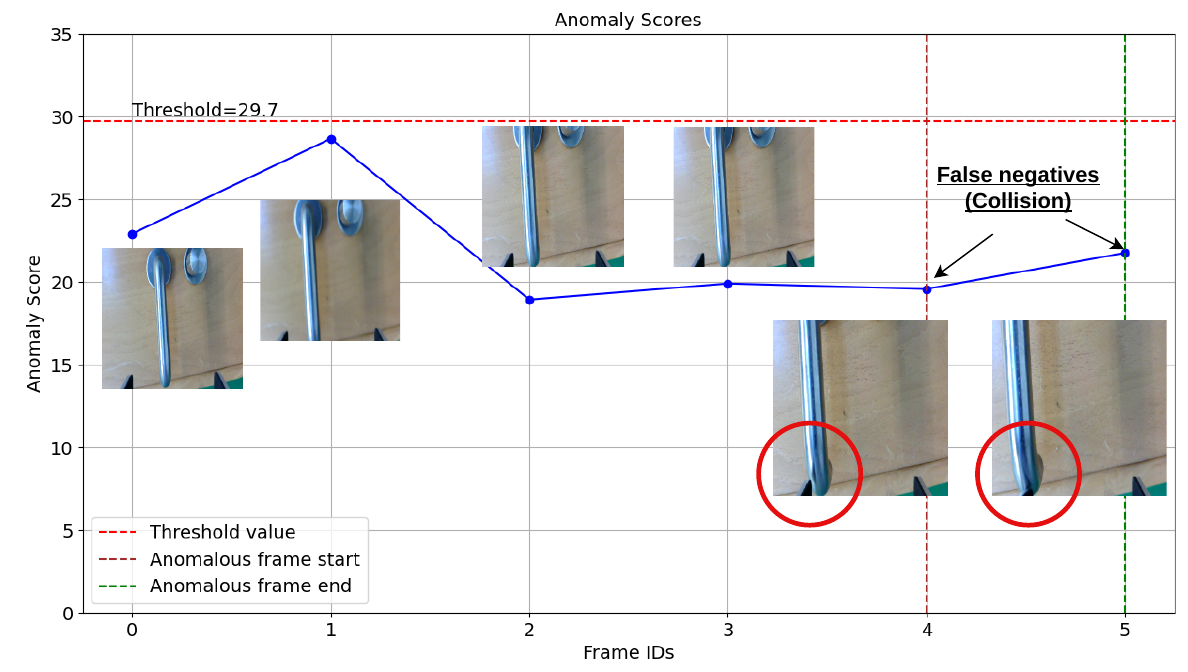}
        \caption{An example of a false negative detection in the door handle reaching task, where a collision of the end effector with the handle is not flagged as an anomaly.}
        \label{fig:anomaly-detection-scores-fn-example}
        \vspace{-0.45cm}
    \end{figure}

    An example of anomaly detection with the trained model during deployment using the threshold value that was found to be optimal\footnote{We use the precision-recall implementation in scikit-learn to find $\tau^*$: \url{https://scikit-learn.org/stable/}} ($\tau^* = 29.7$) is illustrated in Fig. \ref{fig:anomaly-detection-scores-example} in the case of a hand obstruction.
    The detector is not able to detect all anomalies, however; one example is shown in Fig. \ref{fig:anomaly-detection-scores-fn-example}, where a collision with the handle is not detected as an anomaly.

    To evaluate the detector quantitatively, we collected and labelled an additional set of $467$ frames during policy execution ($96$ anomalous and $371$ nominal).
    For this data, the detector with the optimal threshold had a precision of $73.8\%$ and recall of $82.3\%$, resulting in an F-score of $77.8\%$.
    The detector's runtime is dependent on the used hardware and the size of $\mathcal{Z}$; for our evaluation, the detector runs at a frequency of $15 \mbox{Hz}$ on a laptop with an NVIDIA RTX 4060 GPU.

    \subsection{Execution Recovery}
    \label{sec:evaluation:recovery}

    For the recovery, with $ER_1$, the robot sleeps for $4$--$5s$; we consider this to be a reasonable waiting time, particularly for external obstructions.
    With $ER_2$, we locally perturb the arm's position with a value of $\sigma_d = 2cm$; a small value was used to prevent the robot from straying far from its current trajectory and introducing other unwanted anomalies, such as collisions.
    With $ER_3$, we reset the robot's initial position; to train $\mathcal{F}$, we collected a dataset of $45$ starting positions from which the execution policy could be executed successfully.\footnote{We use the implementation in scikit-learn for learning a GMM model.}
    Here, we fit multiple GMMs and choose the number of components that minimises the Bayesian Information Criterion (BIC) \cite{schwarz1978}; for the collected data, a single GMM component was found to be optimal.

    To quantitatively evaluate our method, we compare two versions of our policy: (i) a baseline policy $\tilde{\pi}_\phi$ without the anomaly detection and recovery, and (ii) the failure-aware policy $\pi_\phi$ with the integrated anomaly detection and recovery.
    For the evaluation, we performed $20$ trials of both policies from various starting positions, introducing various anomalies during the process, just as above.
    The results of the evaluation are shown in Tab. \ref{tab:policy-with-recovery}.
    \begin{table}[t]
        \caption{Success rate of the policy in cases where anomalies where introduced (out of $20$ trials)}
        \label{tab:policy-with-recovery}
        \begin{tabular}{L{0.7\linewidth} | R{0.2\linewidth}}
            \cellcolor{gray!10}\textbf{Evaluation mode} & \cellcolor{gray!10}\textbf{Success rate} \\\hline
            Policy without anomaly detection and recovery ($\tilde{\pi}_\phi$) &  $0\%$ \\\hline
            Failure-aware policy ($\pi_\phi$)                                  & $\mathbf{75\%}$ \\\hline
        \end{tabular}
        \vspace{-0.25cm}
    \end{table}
    As can be seen here, $\tilde{\pi}_\phi$ was unable to succeed in any of the trials where anomalies were introduced, while $\pi_\phi$ was able to recover from a large number of anomalies and succeeded in most of the trials.
    A few of these trials failed as well, mostly due to false negative detections that led to collisions with the handle.

    In the trials with $\pi_\phi$, we also evaluate the effectiveness of the individual recovery strategies.
    The results of this evaluation are shown in Tab. \ref{tab:recovery-strategies-door-opening}.
    \begin{table}[t]
        \caption{Number of occurrences of the individual recovery strategies during the failure-aware policy evaluation in the door handle reaching task (over all 20 trials)}
        \label{tab:recovery-strategies-door-opening}
        \begin{tabular}{L{0.3\linewidth} | M{0.29\linewidth} | M{0.27\linewidth}}
            \cellcolor{gray!10}\textbf{Strategy} & \cellcolor{gray!10}\textbf{Number of attempts} & \cellcolor{gray!10}\textbf{Strategy successes} \\\hline
            Pausing ($ER_1$)      & 48 & 3 \\\hline
            Perturbation ($ER_2$) & 45 & 3 \\\hline
            Sampling ($ER_3$)     & 42 & 9 \\\hline
        \end{tabular}
        \vspace{-0.35cm}
    \end{table}
    Here, we see that, in most cases that were recognised as anomalies, $ER_1$ and $ER_2$ were insufficient for recovery, and the robot executed $ER_3$, which was successful in significantly more cases.
    It should be mentioned that the recovery was sometimes triggered by false positive detections, which explains the large number of recovery attempts; we particularly observed false positives due to blurry frames that were caused by the robot's motion.

    \subsection{Generalisation to a Pretrained Policy Model}
    \label{sec:evaluation:generalisationToFoundationModel}

\begin{figure}[t]
        \centering
        \includegraphics[width=0.85\linewidth]{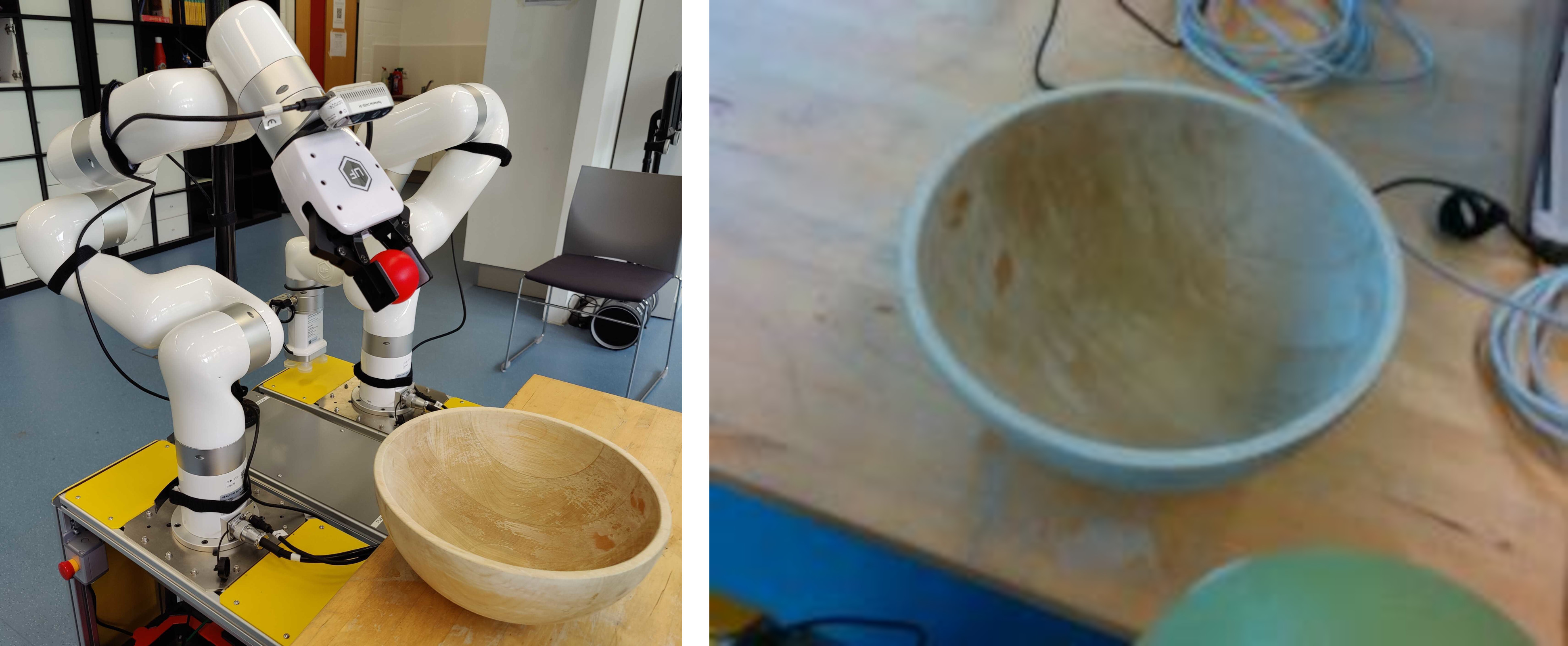}
        \caption{Experiment setup of the object placing task (left) and typical wrist camera view (right)}
        \label{fig:experiment-setup-xarm}
        \vspace{-0.25cm}
    \end{figure}

    To show the policy-agnostic nature of our method, we apply it to a scenario in which a robot uses a pretrained general-purpose policy $\pi_\phi^G$ to put a spherical, apple-like object into a bowl on a table in front of the robot; at the start of each trial, the robot is already holding the object.
    We work with a bimanual mobile manipulator for this task, although only one of the arms is used.
    The setup is illustrated in Fig. \ref{fig:experiment-setup-xarm}.
    For execution, the robot uses a pretrained \emph{Octo-Base} model, with a wrist camera image as state input and the prompt ``You are holding an apple in your hand. Put the apple in the bowl.'' as a goal prompt.
    To ensure the stability of the gripper command provided by the policy, we use a majority filter on the five latest commands, namely we open the gripper if at least three of those indicate that it should be opened.\footnote{The filter parameters were determined empirically.}
    We train the anomaly detection feature extractor using $2{,}440$ nominal images, and determine $\tau^* = 50.0$ using a validation set of $470$ images.

    We evaluate the success rate under three execution conditions (25 trials each): (i) under nominal conditions, (ii) in the presence of anomalies, but without using our method, and (iii) in anomalous conditions using our method.
    In the latter two cases, we manually introduce a variety of anomalies, such as covering the camera or the bowl, removing the bowl while the robot is moving towards it, turning the bowl upside down, or shaking the bowl while the robot is moving towards it.
    The anomalies are introduced either directly at the start of the execution or during it, and are removed after $5-8s$.
    For $\mathcal{F}$, we fit a GMM with three components using the successful starting positions under nominal conditions.

    The results of this evaluation are shown in Tab. \ref{tab:foundation-model-success}.
    \begin{table}[t]
        \caption{Success rate of the pretrained policy on the task of putting an object in a bowl (out of $25$ executions)}
        \label{tab:foundation-model-success}
        \begin{tabular}{L{0.8\linewidth} | R{0.1\linewidth}}
            \cellcolor{gray!10}\textbf{Nominal execution (without anomalies)}                    & $56\%$ \\\hline
            \cellcolor{gray!10}\textbf{Execution in anomalous cases w/o detection and recovery}  & $48\%$ \\\hline
            \cellcolor{gray!10}\textbf{Execution in anomalous cases with detection and recovery} & $\mathbf{60\%}$
        \end{tabular}
        \vspace{-0.45cm}
    \end{table}
    As can be seen, the nominal, zero-shot success rate of the policy is slightly higher than $50\%$; most of the failures in the nominal execution are due to the model releasing the object before reaching the bowl or due to the object bouncing off of the bowl's edge.
    When anomalies are introduced in the execution (and our method is not used), the success rate drops below $50\%$, although it was observed that the policy itself has some recovery ability, presumably due to emergent behaviour (typically manifested by the robot moving slowly while an anomaly was present).
    When using our method, the success rate in the anomalous cases increases to $60\%$, which is comparable with the execution success of the nominal policy (given that the anomalies are eventually removed, the remaining failures are due to the nominal policy).
    This confirms that our method can enable a robot to successfully handle unforeseen anomalies, even though eventually the robot may fail due to inherent limitations of the policy itself.
    In particular, the recovery behaviours were invoked in almost all anomalous cases, except when the bowl was moved to introduce a motion blur; this was seemingly not enough of a disturbance to be detected as an anomaly.

    Tab. \ref{tab:recovery-strategies-bowl-putting} shows the number of times the individual recovery strategies were invoked, and how often the execution was successful after their application.
    \begin{table}[t]
        \caption{Number of occurrences of the individual recovery strategies during the failure-aware policy evaluation in the task of putting an object in a bowl (over all 25 trials)}
        \label{tab:recovery-strategies-bowl-putting}
        \begin{tabular}{L{0.3\linewidth} | M{0.29\linewidth} | M{0.27\linewidth}}
            \cellcolor{gray!10}\textbf{Strategy} & \cellcolor{gray!10}\textbf{Number of attempts} & \cellcolor{gray!10}\textbf{Strategy successes} \\\hline
            Pausing ($ER_1$)      & 37 & 3 \\\hline
            Perturbation ($ER_2$) & 31 & 2 \\\hline
            Sampling ($ER_3$)     & 29 & 7 \\\hline
        \end{tabular}
        \vspace{-4mm}
    \end{table}
    These results are consistent with those in the first task, namely the recovery behaviours are sometimes triggered due to false positive detections, and $ER_3$ is the most successful recovery behaviour.

    \section{DISCUSSION AND CONCLUSION}
    \label{sec:discussion}

    In this paper, we proposed a method that integrates a learned robot policy with visual anomaly detection and subsequent recovery.
    Here, data from nominal policy executions is used to fine-tune a self-supervised DINO-based image feature extractor.
    Anomalies are then detected based on the distance to the nearest neighbour in the dataset of nominal execution features, where the detection threshold is optimised using a dataset of nominal and anomalous images.
    Anomalies trigger a recovery process that involves the sequential execution of different recovery behaviours (pausing the execution, locally perturbing the robot's state, and resetting the robot's state and retrying the execution); for resetting, states are sampled from a success model (in the form of a GMM) that is learned from nominal executions.
    We evaluated our method on the task of reaching a door handle using a dedicated policy (with sim-to-real transfer) and the task of placing an object in a bowl using a pretrained general-purpose language-conditioned policy.
    The results demonstrate that integrating anomaly detection and recovery is important for increasing the robustness of a learned policy, particularly when external anomalies are introduced, although the robustness of the nominal policy itself is just as important for overall success.

    We have various ideas for future work.
    To simplify the evaluation procedure using the dedicated policy, we manually determined if the execution should terminate, but using a complete skill model with initiation and termination classifiers \cite{konidaris2018} is necessary for integration of the policy in long-horizon tasks \cite{sliwowski2025}.
    For training this policy, we used a PyBullet-based simulation due to the simplicity of customisation, but using a photorealistic environment as in \cite{robocasa2024} can improve the sim-to-real policy transfer.
    In this work, we only focused on visual anomalies, as visual data is a rich source of information and visual feature extraction models can be considered to be quite mature; for detecting a larger class of anomalies that are not effectively represented by visual data, for instance collisions as in \cite{sharma2023}, it would be useful to integrate multimodal data, such as force measurements or audio, similar to \cite{inceoglu2021,thoduka2024}.
    Related to this, in applications with concrete requirements on the robustness of the anomaly detector, it may be desirable to replace our detector with one that complies to those requirements more precisely; this would not compromise the operation of our system.
    Finally, as our results show, resetting the execution seems to be the most successful recovery strategy, but, for recovery planning, it would be useful to study the strategies in isolation and identify cases in which each of them is particularly effective.


\addtolength{\textheight}{-4cm}   


\section*{ACKNOWLEDGMENT}

This work was supported in part by the b-it foundation and in part by a starting research grant for Alex Mitrevski provided by Bonn-Rhein-Sieg University of Applied Sciences (project KEROL).

\bibliographystyle{IEEEtran}
\bibliography{references}

\end{document}